# SolarisNet: A Deep Regression Network for Solar Radiation Prediction

Subhadip Dey*
*Indian Institute of Technology Kharagpur*
Kharagpur, WB, India
sdey23@iitkgp.ac.in

Sawon Pratiher*
*Indian Institute of Technology Kharagpur*
Kharagpur, WB, India
sawon1234@gmail.com

Saon Banerjee
*Bidhan Chandra Krishi Viswavidyalaya*
Nadia, WB, India
sbaner2000@yahoo.com

Chanchal Kumar Mukherjee
*Indian Institute of Technology Kharagpur*
Kharagpur, WB, India
ckm@agfe.iitkgp.ernet.in

*Abstract—* **Effective utilization of photovoltaic (PV) plants requires weather variability robust global solar radiation (GSR) forecasting models. Random weather turbulence coupled with assumptions of clear sky model as suggested by Hottel pose significant challenges to parametric & non-parametric models in GSR conversion rate estimation. In addition, a decent GSR estimate requires costly high-tech radiometer and expert dependent instrument handling and measurements, which are subjective in nature. As such, a computer aided monitoring (CAM) system to evaluate PV plant operation feasibility by employing smart grid past data analytics and machine learning is developed. Our algorithm, SolarisNet is a 6-layer deep neural network, which is trained and tested on data collected at two weather stations located near Kalyani metrological site, West Bengal, India. The daily GSR prediction performance using SolarisNet outperforms the existing state of the art, while its efficacy in inferring past GSR data insights to comprehend daily and seasonal GSR variability along with its competence for short term forecasting is discussed.**

*Keywords— Deep learning; Gaussian process regression (GPR); Global solar radiation (GSR); forecasting; time series;*

## I. INTRODUCTION

Kyoto Protocol (KP) like strategic agreements on energy resources reflects the need for long run forecasting of renewable energy time series fluctuations and mitigate the problems of environment degradation due to emission exhausts from non-renewable resources [1]. Photovoltaic systems for industrial and domestic uses require the distribution of grid connected power systems with solar radiation as the main energy source. However direct conversion of solar to electrical energy is costly and has relatively low efficiency [2]. Coupled with grid stability issues concerning scheduling and assets optimization for short-term (monthly)and long-term (yearly) forecasting requires guaranteed knowledge of solar radiation instabilities at local weather stations. All this information is based on satellite observations and data from ground stations, with uncertainty in geographic and time availability of data, and data sampling rate posing significant forecast granularity. To assess the PV plant operation dependability on global solar radiation (GSR), good measurement of GSR using a high class radiometer and correct controlling of the instrument through correct maintenance policy is essential. In order to reduce these involving costs, a solution envisaging data collected by public weather station installed near the plant but in a different location is helpful. In developing countries like India, the use of data analytics based GSR forecasting is attractive because weather station belongs public network bodies and the data are certified and free of cost. Several solar radiation prediction models [3-7], from the available meteorological parameters have been developed [3] from time to time, which includes both parametric and nonparametric algorithms. Razmjoo et al [9], applied the Angstrom-Prescott (AP) methodology to estimate the monthly global solar radiation in Ahvaz and Abadan cites, Iran. Artificial Neural Network (ANN) with Multiple Regression Models (MLR) is compared by Kumar et al. [12-14, 17-18] for monthly GSR prediction and it shows the superiority of ANN models over MLR where the mean absolute percent error (MAPE) is lower in case of ANN. Mohanty et al. [8] review on models categorization based on empirical and soft computing classes highlights the precision advantages of soft computing model in predicted error minimization. A lot of research has been done in this aspect of empirical and soft computing models [3-5]to predict horizontal daily and monthly solar radiation. In this contribution, a deep neural network, named SolarisNet which is particularly suited for GSR prediction is proposed for condition monitoring with minimal set of meteorological parameters like maximum temperature, minimum temperature and sunshine hour data collected at Kalyani public weather stations located West Bengal, India. The comparative evaluation of SolarisNet with existing different parametric and non-parametric models is done to validate the adequacy of SolarisNet in GSR estimation.

## II. RELATED THEORY

### A. Deep Neural Net for GSR Prediction: SolarisNet

*1) Heuristic behind SolarisNet*

The concept of deep convolutional network came from the biological connections in human body [16]. Every body part does not sense the same, some of them sense touch whereas some sense flavor, aroma etc. If we consider these as features of a particular specimen, then the features do not mix altogether at the beginning of the journey from the sensory organ to brain. Rather these all features travel a short length with non-linear modification and before going to decision lobe these mix altogether which discloses the property of the specimen and human take decision. In this short run, these features cross many nodes and internodes of neurons. So, we assumed that in these crossing these features independently acquires non-linearity in them. After a certain length of path, these features start mixing themselves and hence the dimensionality gets reduced which is denoted as nonlinear embedding node. These mixed features again mix with each other to form a decision.

*Equal Contribution.

*2) SolarisNet architecture*

The structure of the proposed deep neural network with three inputs is shown in Fig. 1. The structural details about SolarisNet is given in Table I. The number of mixed layer nodes should be lesser than input nodes to achieve dimensionality reduction and uniform mixture. In this network we have used hyperbolic tangent transfer (tansig) activation function in the first deep hidden layer. From the Non-linear embedding node to second hidden layer log-sigmoid (logsig) activation function is used to increase the correlation within the target data. The following activation functions are used in the SolarisNet layers and are given below:

$$Logsig(n) = \frac{1}{1+e^{-n}}, \text{ and, } tansig(n) = \frac{2}{1+e^{-2n}} - 1$$

The training of the network is done by Levenberg-Marquardt (LM) back propagation technique [7]. LM algorithm uses second order training speed. For the sum of squares performance function the Hessian matrix can be approximated as $H = J^T J$ and the gradient can be computed as $g = J^T e$, where $J$ is the Jacobean matrix that contains first derivatives of the network errors with respect to the weights and biases, and $e$ is a vector of network errors. The Levenberg-Marquardt algorithm uses this approximation to the Hessian matrix in the following Newton-like update [16]:

$$X_{k+1} = X_k [J^T J + \mu I]^{-1} J^T e$$

$\mu$ is decreased after each step and increased whenever there is an increase in the performance function.

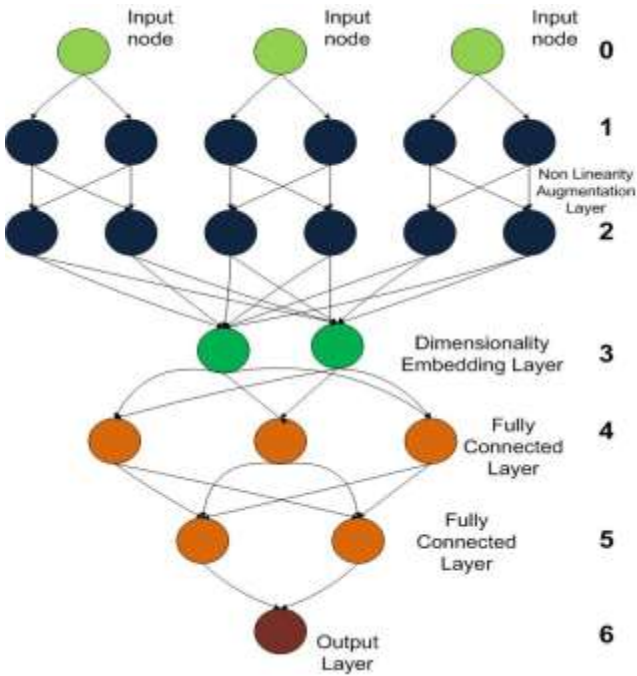

Fig. 1. Deep convolutional regression network.

TABLE-I: Details of SolarisNet architecture

| Layers | Layer type | No of neurons | Activation function |
|---|---|---|---|
| 0-1 | Input | 1x3 | Direct connection |
| 1-2 | Non-linearity augmentation | 2x2x3 | tan sigmoid |
| 2-3 | Dimensionality Embedding | 1x2 | log sigmoid |
| 3-4 | Fully connected | 3 | log sigmoid |
| 4-5 | Fully connected | 2 | log sigmoid |
| 5-6 | Output | 1 | Direct connection |

## III. GAUSSIAN PROCESS REGRESSION (GPR) MODEL

Standard regression model involves training a data set, $\mathcal{D} = \{(x_i, y_i), I = 1, 2 \ldots n\}$, where, $x_i$ and $y_I$ represents the input and output vectors. In the GPR model, it is assumed that the noise is additive, independent and of Gaussian type. Hence, the regression formulation becomes, $y = f(x) + \varepsilon$, where $f(x)$ is some unknown latent function operated element wise on the inputs. $\varepsilon = \{\varepsilon_n\} \in R^N$, is a constant power Gaussian noise given by, $\varepsilon_n \sim \mathcal{N}(0, \sigma_n^2)$ [10, 11]. The GPR assigns a Gaussian process (GP) prior to the unknown function, f (.), instead of parameterizing it [7, 8], and is represented by $f(x) \sim \mathcal{GP}(\mu_x, k(x, \tilde{x}))$. Here, $\mu_x$ and $k(x, \tilde{x})$ signifies the mean and the covariance functions respectively. Moreover, GPR is a better choice when underlying model parameters are unknown. With GP being represented as a set of random variables, it is formulated as:

$$\begin{bmatrix} y \\ y_* \end{bmatrix} \sim \mathcal{N}\left(0, \begin{bmatrix} \mathcal{C}(x,x) + \sigma_n^2 I & \mathcal{C}(x,x_*) \\ \mathcal{C}(x_*,x) & \mathcal{C}(x_*,x_*) \end{bmatrix}\right)$$

Where, $x_*$ is the test data and $\mathcal{C} \in R^{N \times N}$ is generated from the squared-exponential covariance function,

$$k(x, \tilde{x}) = \sigma_f^2 \exp\left(-\frac{|x-\tilde{x}|^2}{2l^2}\right)$$

Here, $l$ is the length-scale, characterizing the distance metric. For short length scales, all estimates are less correlated and the predictive variance increases rapidly away from the data points. The vectors, $y$ and $y_*$ are conditionally distributed as, $y_*|y \sim \mathcal{N}(\mu_f, cov(f_*))$

Where, $\mu_f = k_*^T (\mathcal{C}(x,x) + \sigma_n^2 I)^{-1} y$ represents the matrix of regression coefficients,

And, $cov(f_*) = \mathcal{C}(x_*,x_*) - k_*(\mathcal{C}(x,x) + \sigma_n^2 I)^{-1} k_*$ is the Schur complement. Hence, $cov(f_*)$ depends only on the inputs, $x \in R^N$. In GP model both the prior and likelihood are Gaussian, i.e., $f|x \sim \mathcal{N}(0, \mathcal{C})$ and $y|f \sim \mathcal{N}(f, \sigma_n^2 I)$. The marginal likelihood is given by,

$$p(y|x) = \int p(y|f,x) p(f|x) df$$

The log marginal likelihood is the sum total of the data-fit term, regularization terms and a normalization constant, and is given by,

$$\log(p(y|x)) = -\frac{1}{2}y^T \tilde{C}^{-1} y - \frac{1}{2}\log(|\tilde{C}|) - \frac{n}{2}\log(2\pi)$$

Where, $\tilde{C} = C(x,x) + \sigma_n^2 I$

More details about the existing parametric models like Angstrom-Prescott and non-parametric methods like GPR, SVR and ANN can be traced from [9-13].

## IV. EXPERIMENTAL RESULTS

### A. Acquisition of Global Solar Radiation (GSR) Field Data and Pre-processing

The incoming GSR time series data was measured using a thermopile pyranometer (Make National Instrument and Calibrated by India Meteorological Department, Pune). The thermopile pyranometer is designed to measure the broadband solar radiation flux density with 180° field view and its measurable wavelength ranges from 0.3 μm to 3 μm. The instrument does not require any power to operate and gives output voltage in mili-volt range, which was converted to radiation flux units ($Wm^{-2}$) by suitable standardisation. The hourly interval measurement throughout the day was cumulated to obtain the daily solar radiation in MJ $m^{-2}$ $day^{-1}$. Table I summarizes the descriptive statistics of the two datasets (DS) used for proposed deep learning based method validation. The GSR range, total number of samples in each dataset and along with mean and variance of each one of the meteorological parameters are given. Dataset-1 (DS1) is taken from meteorological site with coordinates 22.97°N 88.43°E while, dataset-2 (DS2) is taken from coordinate's 22.52°N 88.33°E. The field GSR data is subjected to low pass filtering and variational mode decomposition [15] based denoising for removing transients and environmental noises.

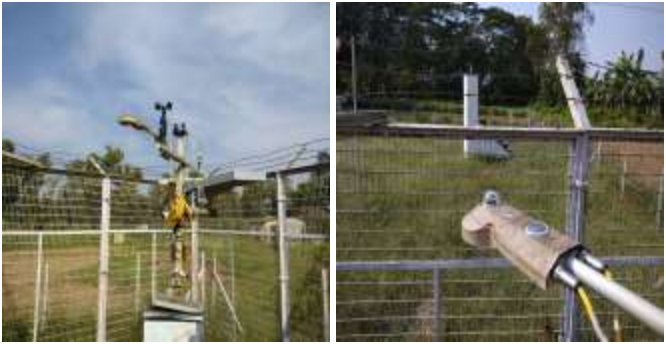

Fig. 2. Field measurement of global solar radiation using pyranometer located at from Kalyani meteorological site, WB, India.

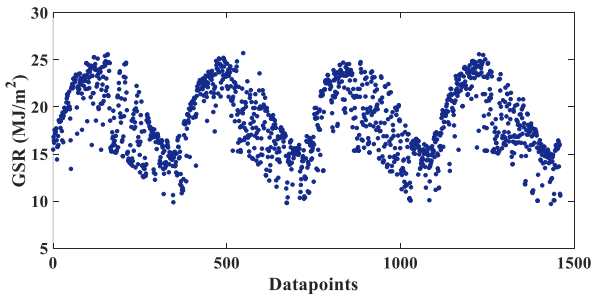

Fig. 3. Daily global solar radiation data pattern.

TABLE-II: DESCRIPTIVE STATISTICS OF DATASET1 AND DATASET 2.

| Dataset (DS) [No. of samples] | | Max. Temp. | | Min. Temp. | | Sunshine hour | |
|---|---|---|---|---|---|---|---|
| | | μ | σ2 | μ | σ2 | μ | σ2 |
| DS1 (1461) | GSR range (MJ/m2) 9.69 to 25.70 | 31.92 | 4.49 | 21.11 | 6.30 | 6.42 | 3.11 |
| DS2 (1064) | GSR range (MJ/m2) 5.60 to 23.50 | 32.23 | 4.41 | 23.29 | 5.08 | 5.13 | 2.98 |

### B. Significance of the considered meteorological parameters

The significance of the three parameters can be estimated from the log length scale vs. length scale number plot. Low log length scale value has high influence to the response of the model. A total number of eleven meteorological parameters (300 observations) from the dataset are given as input in GPR model which discloses length scale number 9 (sunshine hour) has the highest sensitivity to predict the GSR. The second, third and fourth highest sensitive length numbers are 3 (Dry Bulb Temperature I), 2 (minimum temperature), 1 (maximum temperature) respectively. As we are considering for the simplicity of the model as well as the requirement of meteorological instruments we are taking length scale numbers 1, 2 and 9 as input meteorological parameters to the GPR model.

The sensitivity (SA) analysis of the data is done to show the relevancy of the input features. The SA of the GPR for the dataset reveals that sunshine hour is the most sensitive feature. It is obvious that if the sunshine hour changes then the incoming solar radiation will change rapidly, hence the GSR will get affected. The effect of whole day GSR is sensed at night. Decrease in daily GSR causes higher min, temp. which is evident for observations made for cloudy and partially-cloudy day. The SA is shown in Fig. 4.

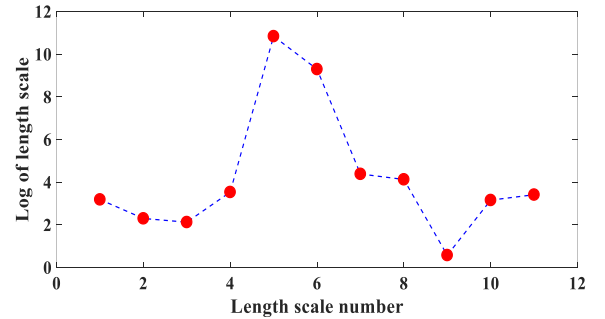

Fig. 4. Log-log plot of the length scale vs. length scale number. A total of eleven meteorological parameters are given as input in the GPR model, out of which parameter 9 (sunshine hour), parameter 3 (Dry bulb temperature), parameter 2 (Minimum temperature) and parameter 1 (maximum temperature) are the most statistical significant parameters to the GPR response.

### C. Training and test data splitting

We have trained different statistical models for the corresponding datasets. In all cases, we standardized the input features. In VMD and GPR combined model the mean was removed from observed GSR data. The data is split in 80% training set and 20% test or holdout set. Hyper-parameters for

GPR model were optimized by maximizing the marginal log-likelihood using the training set.

*D. Evaluation Metrics: Statistical Measures*

Performance assessment of GSR forecasting models are evaluated using four statistical measures which are defined as:

1) *Root mean square error (RMSE) is defined as*
$$\text{RMSE} = \left[\sum_n (H_{pred} - H_{actual})^2 / n\right]^{0.5}$$

2) *Mean absolute error (MAE) is defined as*
$$\text{MAE} = \frac{1}{n}\sum_n |H_{actual} - H_{pred}|$$

3) *Pearson correlation ($\rho$) is defined as*
$$\rho = \frac{\sum_n (H_{pred} - \overline{H}_{pred})(H_{actual} - \overline{H}_{actual})}{\sqrt{\sum_n (H_{pred} - \overline{H}_{pred})^2}\sqrt{\sum_n (H_{actual} - \overline{H}_{actual})^2}}$$

4) *Mean bias error (MBE) is defined as*
$$\text{MBE} = \sum_n (H_{pred} - H_{actual})/n$$

Where, $H_{pred}$ and $H_{actual}$ are the predicted and the actual values of GSR, $\overline{H}_{pred}$ and $\overline{H}_{actual}$ are the average values of predicted and actual GSR. In general a low RMSE is desirable, a positive MBE shows over estimation while a negative MBE indicates under estimation. Best results of each model are shown in Table III. In terms of the statistical measures like mean absolute error (MAE), Pearson correlation coefficient ($\rho$), mean bias error (MBE) are compared for each model.

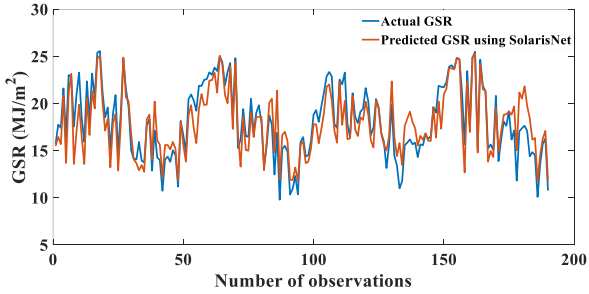

Fig. 5. Representative plots for predicted GSR using SolarisNet.

## V. DISCUSSION

We developed our deep neural net, SolarisNet in MATLABR2016a and have trained it on AMD FX-8320 Octa-core system with 3.50 GHz processor and 16GB RAM. Some of the heavy-duty processes were written in C MEX file to speed up calculation. Moreover, the proposed net requires 1-D convolutions, and therefore implementation wise economical and requires simple hardware. Our net delivered an RMSE of 1.7661 and 1.0492 on two dataset used, which is the best as compared to the state of art. Table III tabulates the performance details.

It took an average of 14.034sec to complete one set of training. Both parametric (Angstrom Eqn.) and non-parametric models (GPR, ANN and SVR) are trained with the field dataset and compared with the proposed SolarisNet based prediction. Also, Table III, illustrates that due to denoising of data, the RMSE got reduced. Fig. 6 depicts the typical prediction plots with prediction error. Fig. 5, illustrates the variation in observed GSR with predicted GSR values for DS1 using SolarisNet utilizing only most significant features. It can be seen that the prediction performance Denoised GPR is better than GPR, SVR and ANN, with the proposed SolarisNet being the best in terms of performance. From the prediction graph, shown in Fig 6. It can be observed that overlapping of predicted GSR is better in SolarisNet whereas the variation of prediction increases for GPR based models if the input test data exceed the limit of training data.

TABLE-III: COMPARISON OF SELECTED STUDIES IN DAILY GLOBAL SOLAR RADIATION PREDICTION.

|  | Method | RMSE | MAE | MBE | $\rho$ |
|---|---|---|---|---|---|
| DS1 | Angstrom Eqn. [9] | 2.6710 | 2.0035 | -1.5431 | 1.1250 |
| DS1 | SVR [12] | 2.7930 | 2.1032 | -1.7618 | 1.0810 |
| DS1 | GPR [10, 11] | 2.3560 | 1.7367 | -1.1626 | 1.1478 |
| DS1 | ANN [13, 14] | 2.0821 | 1.7855 | -0.0224 | 1.2909 |
| DS1 | **Proposed SolarisNet** | **1.7661** | **1.7249** | **-0.3682** | **1.2376** |
| DS2 | Angstrom Eqn.[9] | 3.4731 | 2.7886 | 1.0342 | 0.8712 |
| DS2 | SVR [12] | 1.5961 | 1.3245 | **-0.0209** | 1.3847 |
| DS2 | GPR [10, 11] | 1.3713 | 1.0966 | -0.0300 | 1.1839 |
| DS2 | ANN [13, 14] | 1.3983 | 1.1191 | -0.2116 | 1.1392 |
| DS2 | **Proposed SolarisNet** | **1.0492** | **0.7868** | 0.1270 | 1.0956 |

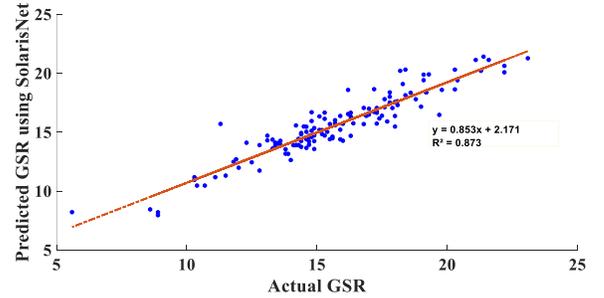

Fig. 6. Observed vs. predicted GSR for DS1 using SolarisNet using most relevant features only.

## VI. CONCLUSION

We develop a deep neural network, SolarisNet for GSR prediction. SolarisNet exceeds the performance of standard existing algorithms in predicting GSR from experimental solar radiation field data. Further, the sensitivity analysis on GPR mean and variance shows the relevancy of minimum temperature and sunshine hour is higher than maximum temperature in GSR forecasting. The presence of non-linear interrelationship among the feature variables is exploited by employing a non-linear mapping layer in the SolarisNet which helps in lowering the RMSE for predicted and measured GSR values. Alternatively, the authors are considering the exploitation of short term temporal and meteorological information in designing long short time networks (LSTN) to improve the robustness of model parameters in estimation.

ACKNOWLEDGMENT

The authors would like to thank Indian Metrological Department(IMD), India for collecting the field data.